\documentclass{article} 
\usepackage{nips14submit_e,times}
\usepackage{hyperref}
\usepackage{url}

\usepackage{algorithm,algorithmic,color}
\usepackage{float}
\usepackage{amssymb}
\usepackage{amsmath}
\usepackage{amsthm}
\usepackage{subfig}
\usepackage{graphicx}
\usepackage{color}
\usepackage{stmaryrd}

\definecolor{darkblue}{rgb}{0,0.05,0.35}
\definecolor{darkgreen}{rgb}{0,0.6,0}

\newtheorem*{mydef}{Definition}

\title{A Structured Variational Auto-encoder\\ for Learning Deep Hierarchies of Sparse Features}

\author{Tim Salimans \\
\texttt{salimanstim@gmail.com} \\
}

\newcommand{\E}{\mathbb{E}}

\DeclareMathOperator{\maximum}{maximum}

\nipsfinalcopy 

\begin{document}

\maketitle

\begin{abstract}
In this note we present a generative model of natural images consisting of a deep hierarchy of layers of latent random variables, each of which follows a new type of distribution that we call \textit{rectified Gaussian}. These rectified Gaussian units allow spike-and-slab type sparsity, while retaining the differentiability necessary for efficient stochastic gradient variational inference. To learn the parameters of the new model, we approximate the posterior of the latent variables with a variational auto-encoder. Rather than making the usual mean-field assumption however, the encoder parameterizes a new type of structured variational approximation that retains the prior dependencies of the generative model. Using this structured posterior approximation, we are able to perform joint training of deep models with many layers of latent random variables, without having to resort to stacking or other layerwise training procedures.
\end{abstract}

\section{A structured variational auto-encoder model}
\label{sec:model}
We propose a directed generative model consisting of a hierarchy of layers of latent features $\mathbf{z}^{0},\mathbf{z}^{1},\ldots,\mathbf{z}^{L}$, where the features in each layer are generated independently conditional on the features in the layer above, i.e.\ $z_{i}^{j} \sim p(z_{i}^{j} | \mathbf{z}^{j-1})$. For the conditional distribution $p()$ we propose what we call the \textit{rectified Gaussian distribution} $RG(\mu_{i}^{j}, \sigma_{i}^{j})$. We can define this distribution by describing how we can sample from it:

\begin{mydef}[\emph{Rectified Gaussian distribution}]
\[\text{If } \epsilon \sim N(0,1), \text{ and } z_{i}^{j} = \maximum( \mu_{i}^{j} + \sigma_{i}^{j}\epsilon, 0) \text{ then } z_{i}^{j} \sim RG(\mu_{i}^{j}, \sigma_{i}^{j}).\]
\end{mydef}

The rectified Gaussian distribution is thus a mixture of a point mass at zero, and a truncated Gaussian distribution with support on the positive real line. Both the mass at zero and the shape of the truncated Gaussian component are determined by the same parameters. Because of this property, the random draw $z_{i}^{j}$ is differentiable in $(\mu_{i}^{j}, \sigma_{i}^{j})$ for fixed $\epsilon$, a property we will exploit later to perform efficient stochastic gradient variational inference.

For the top layer of latent features $\mathbf{z}^{0}$, we define $\mu^{0}$ to be a learnable parameter vector. The standard deviations $\sigma_{i}^{0}$ of the top layer are fixed at 1. After that, the parameters of each layer are recursively set to be
\[\mu^{j} = \mathbf{b}_{\mu}^{j} + W_{\mu}^{j} \cdot \mathbf{z}^{j-1}, \hspace{1cm} \sigma^{j} = \exp\left( \mathbf{b}_{\sigma}^{j} + W_{\sigma}^{j} \cdot \mathbf{z}^{j-1} \right),\]
where $\mathbf{b}_{\mu}^{j}$ and $\mathbf{b}_{\sigma}^{j}$ are (column) parameter vectors, $W_{\mu}^{j}$ and $W_{\sigma}^{j}$ are parameter matrices, and $\cdot$ defines the matrix-vector dot product. The exponential function $\exp()$ is applied elementwise.

After generating the last layer of latent features $\mathbf{z}^{L}$, we generate the observed data $\mathbf{x}$ from an appropriate conditional distribution $p(\mathbf{x} | \mathbf{z}^{L})$. For example, for binary data we use independent Bernoulli distributions, where the probabilities are given by applying the logistic sigmoid to another linear transformation of the latent features $\mathbf{z}^{L}$. For continuous data we could use independent Gaussian distributions with a similar parameterization to the rectified Gaussian units used for the latent features. By marginalizing out the latent features, we now end up with a generative model $p_{\theta}(\mathbf{x})$ for the observed data, where we use $\theta$ to denote all model parameters.

\textbf{Encoder and variational posterior approximation}\\
In order to learn the parameters $\theta$ of our generative model $p_{\theta}(\mathbf{x})$, we optimize a variational lower bound on the log marginal likelihood. For $n$ data vectors $\mathbf{x}^{1},\mathbf{x}^{2},\ldots,\mathbf{x}^{n}$ this lower bound is given by
\begin{equation}
\label{eq:lower_bound}
\sum_{k=1}^{n} \log p_{\theta}(\mathbf{x}_{k}) \geq L(\theta) = \sum_{k=1}^{n} \E_{q_{\psi}(\mathbf{z}_{k}|\mathbf{x}_{k})}[ \log p_{\theta}(\mathbf{z}_{k},\mathbf{x}_{k}) - \log q_{\psi}(\mathbf{z}_{k}|\mathbf{x}_{k}) ],
\end{equation}
where we have introduced a variational distribution $q_{\psi}(\mathbf{z}_{k}|\mathbf{x}_{k})$ over all layers of latent features $\mathbf{z}_{k}$ for data vector $k$. By optimizing the variational lower bound with respect to the parameters $\psi$ of this variational distribution, we can fit $q_{\psi}()$ to the posterior distribution of the latents $p_{\theta}(\mathbf{z}_{k}|\mathbf{x}_{k})$. If we jointly maximize with respect to the parameters of the generative model $\theta$ we are performing an approximate form of maximum likelihood learning.

Following the principle of \textit{variational auto-encoding} \cite{kingma2013auto, rezende2014stochastic}, we define the variational posterior approximation $q_{\psi}()$ to be a parameterized function of the data. Rather than using a factorized mean-field posterior approximation however, we use the parameterized encoder to define a \textit{structured} posterior approximation. Like \cite[section 7.1]{salimans2013fixed} we choose the structure of the variational distribution $q()$ to mirror that of the generative model. That is we choose $q_{\psi}(\mathbf{z}|\mathbf{x}) = q_{\psi}(\mathbf{z}^{0}|\mathbf{x})q_{\psi}(\mathbf{z}^{1}|\mathbf{x},\mathbf{z}^{0}) \ldots q_{\psi}(\mathbf{z}^{L}|\mathbf{x},\mathbf{z}^{L-1})$ to have exactly the same structure as the prior $p(\mathbf{z})$. Each of the conditionals $q_{\psi}(\mathbf{z}^{l}|\mathbf{x},\mathbf{z}^{l-1})$ are once again Rectified Gaussian, with independent marginals $RG(\hat{\mu}_{i}^{j}, \hat{\sigma}_{i}^{j})$, but now with different parameters $(\hat{\mu}_{i}^{j}, \hat{\sigma}_{i}^{j})$ that we allow to depend on the data. We define these parameters as follows:
\[\hat{\sigma}_{i}^{j} = [(\sigma_{i}^{j})^{-2} + (\tilde{\sigma}_{i}^{j})^{-2}]^{-1/2}, \hspace{1cm} \hat{\mu}_{i}^{j} = [(\sigma_{i}^{j})^{-2}\mu_{i}^{j} + (\tilde{\sigma}_{i}^{j})^{-2}\tilde{\mu}_{i}^{j}](\hat{\sigma}_{i}^{j})^{2},\]
where $(\mu_{i}^{j}, \sigma_{i}^{j})$ are the parameters as defined for the prior, and $(\tilde{\mu}_{i}^{j}, \tilde{\sigma}_{i}^{j})$ can be interpreted as the parameters of an \textit{approximate Gaussian likelihood term} that is applied to the latent features $z_{i}^{j}$ before rectification. The parameters of this approximate likelihood term are computed recursively, mirroring the parameters of the prior, but now bottom-up rather than top-down.
\[\tilde{\mu}^{j} = \tilde{\mathbf{b}}_{\mu}^{j} + \tilde{W}_{\mu}^{j} \cdot \maximum(\tilde{\mu}^{j+1},\mathbf{0}), \hspace{1cm} \tilde{\sigma}^{j} = \exp\left( \tilde{\mathbf{b}}_{\sigma}^{j} + \tilde{W}_{\sigma}^{j} \cdot \maximum(\tilde{\mu}^{j+1},\mathbf{0}) \right),\]
where $\maximum(\tilde{\mu}^{j+1},\mathbf{0})$ denotes elementwise rectification of the \textit{approximate likelihood means} $\tilde{\mu}^{j+1}$ in the level below (i.e.\ the previous level in the hierachy when moving up from the data).

\textbf{Optimization of the variational lower bound}\\
After having defined our generative model $p()$ and approximate posterior $q()$, we can evaluate the variational lower bound \eqref{eq:lower_bound} and optimize it to fit our model to the data. We evaluate the variational lower bound by first doing a fully deterministic upward pass up the hierarchy with our encoder model, starting at the data. We then perform a downward pass, where at each level we analytically calculate the KL-divergence between the approximate posterior term $q_{\psi}(\mathbf{z}^{l}|\mathbf{x},\mathbf{z}^{l-1})$ and prior $p_{\theta}(\mathbf{z}^{l}|\mathbf{z}^{l-1})$. This divergence is given by
\[
D_{KL}[q_{\psi}(\mathbf{z}^{l}|\mathbf{x},\mathbf{z}^{l-1}) | p_{\theta}(\mathbf{z}^{l}|\mathbf{z}^{l-1})] = \sum_{i=1}^{n} D_{KL}[q_{\psi}(z^{l}_{i}|\mathbf{x},\mathbf{z}^{l-1}) | p_{\theta}(z^{l}_{i}|\mathbf{z}^{l-1})],
\]
since both the conditional prior and approximate posterior are fully factorized. The KL divergence between the elements of $\mathbf{z}^{l}$ is given by
\begin{eqnarray}
D_{KL}[q_{\psi}(z^{l}_{i}|\mathbf{x},\mathbf{z}^{l-1}) | p_{\theta}(z^{l}_{i}|\mathbf{z}^{l-1})] = Q_{\psi}(z^{l}_{i}|\mathbf{x},\mathbf{z}^{l-1})\log[Q_{\psi}(z^{l}_{i}|\mathbf{x},\mathbf{z}^{l-1})/P_{\theta}(z^{l}_{i}|\mathbf{z}^{l-1})] \nonumber\\
+ [1-Q_{\psi}(z^{l}_{i}|\mathbf{x},\mathbf{z}^{l-1})]\E_{q_{\psi}(z^{l}_{i}|z^{l}_{i}>0,\mathbf{x},\mathbf{z}^{l-1})}\log[q_{\psi}(z^{l}_{i}|\mathbf{x},\mathbf{z}^{l-1})/p_{\theta}(z^{l}_{i}|\mathbf{z}^{l-1})],
\label{eq:elem_kl}
\end{eqnarray}
where $Q,P$ denote the CDFs corresponding to the Gaussian distribution of $z^{l}_{i}$ before rectification. All terms in \eqref{eq:elem_kl} can be computed analytically using univariate Gaussian integrals.

After calculating the KL divergence between $q_{\psi}(\mathbf{z}^{l}|\mathbf{x},\mathbf{z}^{l-1})$ and $p_{\theta}(\mathbf{z}^{l}|\mathbf{z}^{l-1})$, we sample $\mathbf{z}^{l}$ from the approximate posterior and proceed to the next level $l+1$. This way we construct an unbiased approximation of the variational lower bound that is continuously differentiable in all parameters $(\psi,\theta)$. We optimize the variational lower bound over these parameters using stochastic gradient descent using Adamax, a variant of the Adam optimization algorithm \cite{kingma2014adam}. During training we also use Batch-Normalization at each level of both the upward and downward passes to help regularize the training objective and to speed up convergence.

\section{Experiments}
\label{sec:experiments}

\textbf{Permutation invariant MNIST}\\
We evaluate our generative model on the binarized MNIST dataset of handwritten digits. Here we use a model with 4 layers of latent features. From top to bottom, the layers have 50, 100, 200, and 300 units per layer: Unlike variational auto-encoders with a single layer of random variables our model does not have to propagate all information to the top level, so we can let the number of units go down as we move up the hierarchy. We use small minibatches of 150 examples to maximize the regularizing effect of Batch Normalization, and we deliberately limit the number of units in our model: Otherwise, the model easily overfits on the training data. After 500 epochs of training the model with these settings we reach a variational lower bound of $-92.5$ nats on the test set, which is comparable to the performance of variational auto-encoders with a single layer of random variables and multiple deterministic layers.

\section{Conclusion}
\label{sec:conclusion}
We have proposed a new kind of variational auto-encoder containing many layers of stochastic latent variables with sparse activations. For specifying our inference network, we followed \cite[section 7.1]{salimans2013fixed} by using a structured top down variational approximation that combines our hierarchical model with conjugate approximate likelihood terms. This combination allows us to perform joint training of deep models with many layers of latent random variables, without having to resort to stacking or other layerwise training procedures. A similar structured variational approximation was recently used in the independently developed probabilistic ladder network of \cite{sonderby2016train}.

Although the ability to jointly train deep models with many layers of stochastic latent variables is very attractive from a theoretical perspective, neither our method nor the method of \cite{sonderby2016train} has so far improved dramatically over simpler models with a single layer of latent variables. Developing deep generative models that more fully utilize the additional expressivity gained by having multiple layers of latent random variables is an important goal for future work.

\bibliographystyle{unsrt}
\bibliography{biball}

\end{document}